\documentclass[letterpaper, 10 pt, conference]{ieeeconf}
\IEEEoverridecommandlockouts
\usepackage{cite}
\usepackage{amsmath,amssymb,amsfonts}
\usepackage{algorithmic}
\usepackage{graphicx}
\usepackage{textcomp}
\usepackage{xcolor}
\usepackage{soul}

\usepackage{tikz,lipsum,lmodern}
\usepackage[most]{tcolorbox}

\usepackage{hyperref}       
\usepackage{url}            

\usepackage{booktabs}       
\usepackage{amsfonts}       
\usepackage{nicefrac}       
\usepackage{microtype}      
\usepackage{xcolor}         

\usepackage{times}
\usepackage{framed}
\usepackage{color}
\definecolor{shadecolor}{rgb}{0.92,0.92,0.92}
\usepackage{epsfig}
\usepackage{amssymb}
\usepackage{mathtools}
\let\labelindent\relax
\usepackage{enumitem}
\usepackage{setspace}
\usepackage{booktabs}
\usepackage{color,xcolor}
\usepackage{bbding}
\usepackage{diagbox}
\usepackage{bm}
\usepackage{multirow}
\usepackage{booktabs}
\usepackage{colortbl}

\def\BibTeX{{\rm B\kern-.05em{\sc i\kern-.025em b}\kern-.08em
    T\kern-.1667em\lower.7ex\hbox{E}\kern-.125emX}}
\begin{document}

\title{Human Demonstrations are Generalizable Knowledge for Robots}

\author{Te Cui$^{1\dagger}$, Tianxing Zhou$^{1, 2\dagger}$, Mengxiao Hu$^1$, Haoyang Lu$^1$, Zicai Peng$^1$,  \\Haizhou Li$^1$, Guangyan Chen$^1$, Meiling Wang$^1$,  Yufeng Yue$^{1*}$ 
\thanks{This work was supported by National Key RD Program of China (Grant No.2024YFB4708900). It was also supported in part by the Natural Science Foundation of China under Grant 62473050, 62233002, 624B2025.}
\thanks{*Corresponding author: Yufeng Yue (yueyufeng@bit.edu.cn)}
\thanks{$^\dagger$ Equal Contribution}
\thanks{$^1$ School of Automation, Beijing Institute of Technology, Beijing, 100081, China}
\thanks{$^2$ Beijing Zhongguancun Academy, Beijing, 100094, China}
}

\maketitle
\thispagestyle{empty}
\pagestyle{empty}

\begin{abstract}
Learning from human demonstrations is an emerging trend for designing intelligent robotic systems.  However, previous methods typically regard videos as instructions, simply dividing videos into action sequences for robotic repetition, which pose obstacles to generalization to diverse tasks or object instances. In this paper, we propose a different perspective, considering human demonstration videos not as mere instructions, but as a source of knowledge for robots. Motivated by this perspective and the remarkable comprehension and generalization capabilities exhibited by large language models (LLMs), we propose DigKnow, a method that DIstills Generalizable KNOWledge with a hierarchical structure. Specifically, DigKnow begins by converting human demonstration video frames into observation knowledge. This knowledge is then subjected to analysis to extract human action knowledge and further distilled into pattern knowledge that comprises task and object instances, resulting in the acquisition of generalizable knowledge with a hierarchical structure. In settings with different tasks or object instances, DigKnow retrieves relevant knowledge for the current task and object instances. Subsequently, the LLM-based planner conducts planning based on the retrieved knowledge, and the policy executes actions in line with the plan to achieve the designated task. Utilizing the retrieved knowledge, we validate and rectify planning and execution outcomes, resulting in a substantial enhancement of the success rate.  Experimental results across a range of tasks and scenes demonstrate the effectiveness of this approach in facilitating real-world robots to accomplish tasks with the knowledge derived from human demonstrations. 

\end{abstract}

\section{Introduction}
The development of deep learning empowers autonomous agents, including robots, to acquire intricate and adaptable behavioral skills suitable for diverse and unstructured environments. In light of the substantial progress in large language models (LLMs) in recent years, many recent studies 
\cite{liu2023reflect,chen2024vlmimic} 
have explored the application of LLMs in the robot learning field, yielding significant improvements. These methods enhance the ability of robots to comprehend and generate natural language, enabling more intelligent and natural conversational capabilities. Moreover, these LLM-based approaches demand only a limited number of reference examples, eliminating the need for extensive training data collection. Nonetheless, these methods primarily concentrate on text-based task planning, facing challenges related to either insufficient provided information or complex input text content requirements, which pose obstacles for end-users to instruct robots effectively.

Learning from human demonstrations \cite{wake2023gpt, shao2021concept2robot,chen2025graphMimic,chen2025fmimic} is a promising method to instruct robots, as it typically does not necessitate specialized knowledge of robotics technology and involves minimal or no textual input for teaching robots new tasks, facilitating the transfer of research-based robot prototypes to real-world applications. This approach enables end-users to conveniently provide specific instructions to robots tailored to their individual needs within their respective environments. Previous methods predominantly train custom models on extensive datasets of robot actions, necessitating substantial data collection efforts. More importantly, these methods typically treat videos as instructions, dividing them into action sequences for straightforward robotic repetition, thereby impeding generalization across a variety of tasks and object instances. 
In this paper, we argue that human demonstration videos are a source of knowledge for robots. 
Motivated by this perspective and the impressive capabilities demonstrated by LLMs, we naturally ask the question: Can the LLMs serve as an effective knowledge learner conditioned on human demonstrations?

\begin{figure}[!t]
\setlength{\abovecaptionskip}{5pt}
    \centering
    \includegraphics[width=8.5cm]{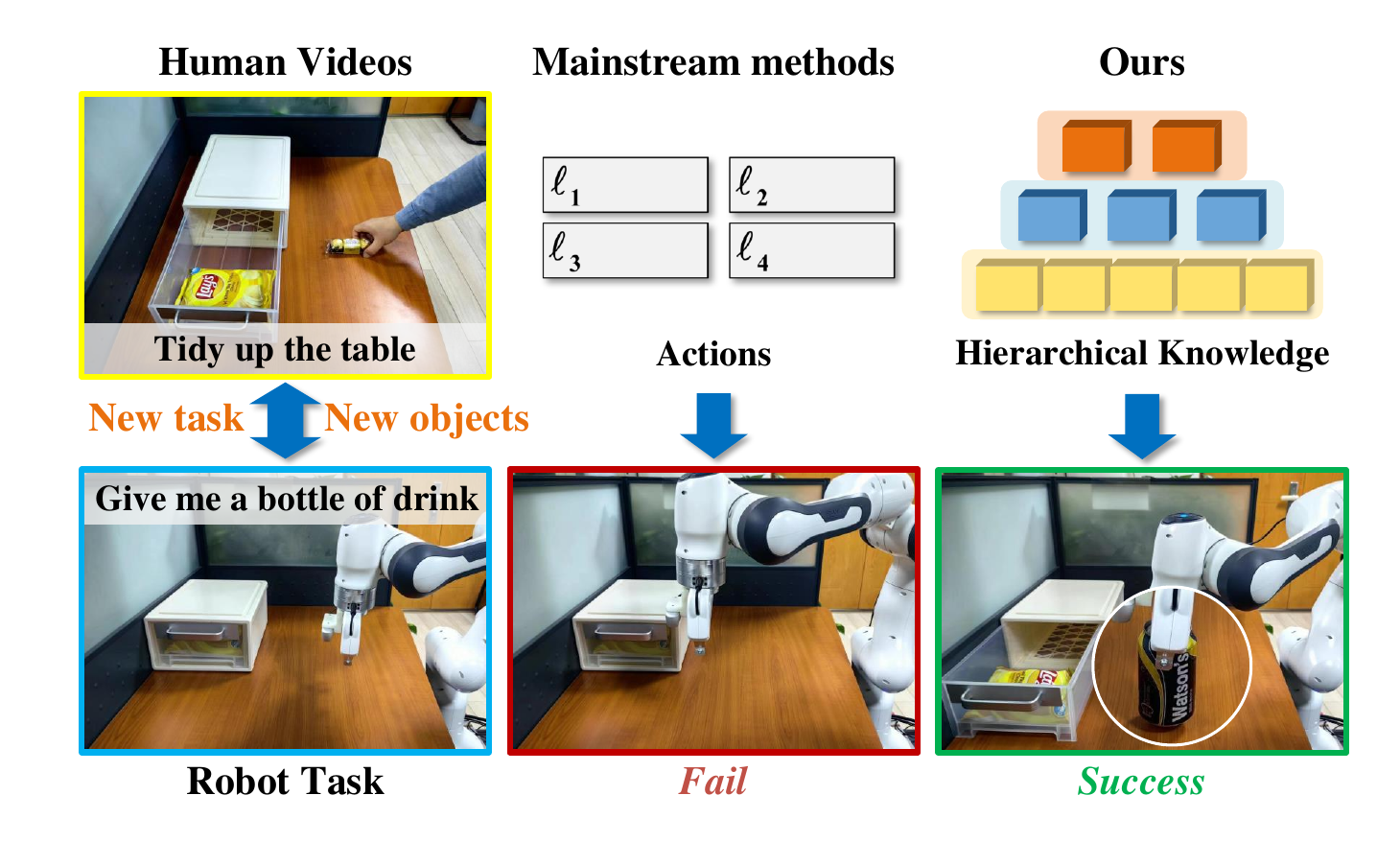}
    \caption{Given a human demonstration video, our DigKnow distills it into hierarchical, generalizable knowledge, rather than mere instructions as employed in prior approaches. In scenarios involving different tasks or object instances, our DigKnow exhibits effective generalization through the utilization of the retrieved knowledge.}
    \label{fig1}
    \vspace{-15pt}
    \end{figure}

To answer this question, we exploit LLMs for robot learning from human demonstration videos. However, it is challenging in terms of analysis, distillation, and generalization due to the following reasons: (\uppercase\expandafter{\romannumeral1}) \textbf{Analysis}. While GPT-4V can be utilized for video analysis and action sequence generation \cite{wake2023gpt}, its limited ability to assess the relative positions of objects hinders its overall accuracy. To address this, we construct scene graphs, convert them into textual descriptions, and subsequently employ LLMs to generate action sequences by comparing consecutive scene graphs. This approach effectively captures relative positional relationships via scene graphs, thereby enhancing the accuracy of video analysis. (\uppercase\expandafter{\romannumeral2}) \textbf{Distillation}. A singular form of knowledge encounters difficulty in addressing both high-level and low-level requirements.  To mitigate this challenge, we distill video analysis into hierarchical knowledge, encompassing observation, action, and pattern levels. For more efficient acquisition of relevant knowledge, we differentiate between task and object knowledge, enabling our method to retrieve knowledge for the current task and objects separately. (\uppercase\expandafter{\romannumeral3}) \textbf{Generalization}. While hierarchical knowledge is utilized to facilitate the generalization of robots to novel scenarios, this capacity is nonetheless constrained by the instability in the outputs of LLMs and the execution of actions. To address this issue, we leverage hierarchical knowledge to further validate and rectify planning and execution outcomes, which effectively enhances the success rate of task completion and strengthens the generalization capability.

Based on the above analysis, we propose a novel LLM-based framework for robot learning from human demonstration videos, called DigKnow. Given a human video, our approach initiates by constructing scene graphs to acquire observation knowledge. We then identify keyframes and employ LLMs to extract action knowledge through a comparative analysis of scene graphs between consecutive keyframes. Subsequently, this knowledge is subsequently distilled into pattern knowledge, with separate consideration for task and object instances.  In scenarios involving different tasks or object instances, our DigKnow retrieves knowledge relevant to the current task and object instances. The LLM-based planner performs planning based on retrieved knowledge, and a low-level policy is employed for execution to accomplish the task. To enhance the generalization ability, we employ retrieved knowledge for additional validation and correction of planning and execution outcomes. As shown in Fig. \ref{fig1}, our DigKnow exhibits robust generalization capabilities, even when confronted with novel tasks and object instances.

Our main contributions can be summarized as follows: 
\begin{itemize}
\setlength{\itemsep}{2pt}
\setlength{\parsep}{2pt}
\setlength{\parskip}{2pt}
\item A novel LLM-based robot learning framework, termed DigKnow, is proposed. DigKnow distills human videos into generalizable knowledge and retrieves relevant knowledge to facilitate generalization in unseen scenarios. 
\item  An efficient video analysis method is proposed, which utilizes LLMs to generate action sequences by comparing scene graphs, achieving accurate video analysis without additional training.
\item A hierarchical knowledge structure based on video is proposed to facilitate the retrieval of relevant knowledge,  thus enabling our method to generalize effectively in different scenarios. 
\item A knowledge-based correcting method is proposed to validate and rectify planning and execution results, thereby improving task completion success rates and bolstering generalization capabilities. 
\end{itemize}

\begin{figure*}[t]
 \setlength{\abovecaptionskip}{-10pt}
    \begin{center}
    \includegraphics[width=18cm]{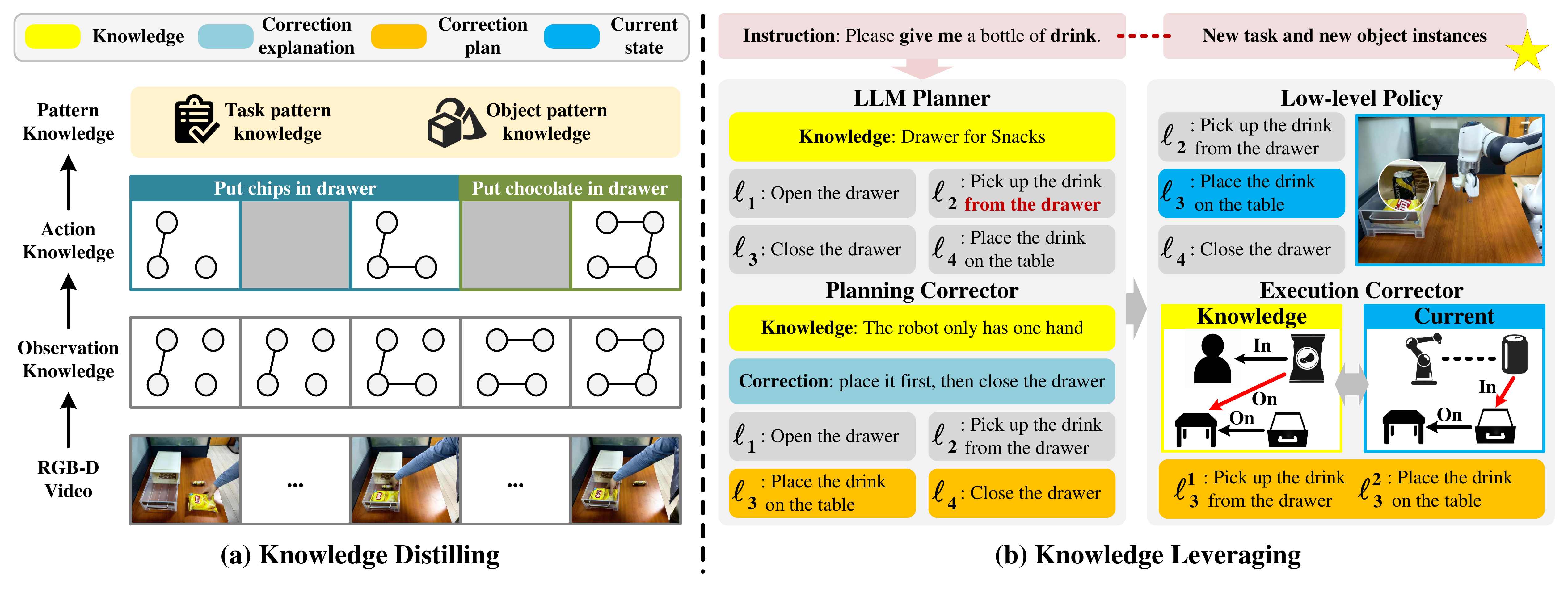}
    \end{center}
       \caption{{ Overall architecture of our DigKnow.}  (a) The human demonstration videos are distilled into hierarchical knowledge, spanning observation, action, and pattern levels. (b) In situations involving unseen  tasks or object instances, the LLM-based planner utilizes the retrieved knowledge for planning, and a low-level policy is employed for execution. Furthermore, we employ retrieved knowledge for further validation and correction of planning and execution outcomes.}
    \label{Figure2}
\vspace{-12pt}
\end{figure*}

\section{Related work}

\vspace{10pt}

\subsection{LLM-based Task Planning}
Large Language Models have demonstrated outstanding performance in various domains, including text comprehension and language translation \cite{llama,touvron2023llama2}. In recent research, there is a growing interest in integrating LLMs with robotic systems to enhance robots' high-level environmental awareness and task comprehension abilities \cite{shridhar2023perceiver,brohan2023rt}. However, most of them \cite{shridhar2023perceiver,brohan2023rt} demand extensive expert-collected robot data and entail the need for data reacquisition and model retraining when applying them to different or expanded robotic contexts. This limitation hinders their adaptability, rendering their direct application to varied tasks and contexts a challenging endeavor. Consequently, researchers have leveraged large language models for task decomposition and motion planning in robotics, translating human instructions into a sequence of precise executable steps \cite{ding2022robottaskplanning,ding2023taskandmotionplaning}, thereby enhancing the versatility of these approaches. SayCan \cite{brohan2023saycan} utilizes LLM to formulate a set of natural language-expressed executable robot actions, concurrently restricting the output space of large language models to enhance the resulting plans. PROGPROMPT \cite{singh2023progprompt} proposes a programmatic LLM prompting structure enabling planning across diverse environments, robot capabilities, and tasks. However, these methods predominantly focus on text-based task planning, encountering difficulties related to either inadequate information provided or intricate input text content requirements, thereby presenting obstacles to effective robot instruction by end-users.

\vspace{10pt}

\subsection{Robot Learning from Human Demonstrations}
Learning from human demonstrations enables robots to quickly acquire the ability to complete specified tasks \cite{schaal1996lfd,ravichandar2020recentlfd}. Previous methodologies \cite{wake2023gpt,shao2021concept2robot,qin2022dexmv} construct datasets of human actions to serve as training data, supervising the robotic motion learning process. Concept2Robot \cite{shao2021concept2robot} employs a two-stage model, trained on an extensive human action dataset, facilitating the robotic arm in executing human instructions. DexMV \cite{qin2022dexmv} utilizes a specialized device to record human hand motions, serving as a reference for model learning and optimizing the motion postures of the robotic arm. However, the processes of collecting and annotating training data can lead to increased time consumption and experimental costs. Furthermore, these methods entail the need for model retraining when applying them to different or expanded robotic contexts. To alleviate these limitations, Microsoft leverages GPT-4(V) to interpret human demonstration videos and formulate motion instructions for robots \cite{wake2023gpt4v}. This approach obviates the need for extensive data collection and model retraining, significantly enhancing system reusability in research. Nonetheless, these methods typically treat videos as instructions, dividing them into action sequences for straightforward robotic repetition, thereby impeding generalization across a variety of tasks and object instances. 

\newpage

\section{DigKnow}

Given the human demonstration video $ \bm{V} \in \mathbb{R}^{T \times 4 \times H \times W}$, representing a sequence of moving images $I$, the DigKnow pipeline, as depicted in Fig. \ref{Figure2}, begins with the construction of scene graphs to capture observation knowledge. Subsequently, DigKnow identifies keyframes and employs LLMs to generate action knowledge by comparatively analyzing consecutive keyframes. This action knowledge, comprising action sequences along with their associated scene graphs, 
is then distilled into pattern knowledge. The resulting hierarchical knowledge is stored in the knowledge base for future retrieval. In situations involving unseen tasks or object instances, our DigKnow retrieves saved knowledge relevant to the current task and object instances. The LLM-based planner utilizes the retrieved knowledge for planning, with a low-level policy employed for execution to accomplish the task. 
Upon completion of planning or action execution, we harness the retrieved hierarchical knowledge for further validation and rectification.

\subsection{Generalizable Knowledge Distillation} \label{Knowledge}
Previous methods typically regard videos as instructions, simply dividing them into action sequences for robotic repetition, presenting difficulties in achieving generalization across various tasks or object instances. 
In this section, we propose to distill human videos into generalizable knowledge with a hierarchical structure. 
The hierarchical knowledge structure contains the observation, action, and pattern levels, with the pattern level separately considering object instance and task patterns. 

\textbf{Observation level}:
To comprehend human-environment interactions, 
the observation level contains inter-object relations, human-object relations, and object state information, represented as scene graphs extracted from downsampled frames.

Given the RGB-D human demonstration video $ \bm{V} \in \mathbb{R}^{T \times 4 \times H \times W}$, we generate the downsampled video $ \bm{\Tilde{V}} \in \mathbb{R}^{T' \times 4 \times H \times W}$  by applying a downsampling scale $d$, where $T' = \frac{T}{d}$.  For each frame $\bm{I}_t$ in the downsampled video, we perform open vocabulary semantic segmentation \cite{kirillov2023segany, liu2023grounding} on the RGB image to obtain the segmentation results $\bm{I}_t^{s}$. 
Subsequently, we project the segmentation results into a 3D semantic point cloud, utilizing the observed depth information. 
From the derived semantic point cloud, we compute spatial relations, encompassing both inter-object and human-object interactions. These relations are based on well-established and commonly used spatial relationship definitions. Additionally, we utilize the VLM to predict the state of each detected object. Consequently, the scene graph $\bm{G}_{t} = \{N, E\}$ is constructed, which describes nodes $N$, including object nodes and an additional human node, along with their spatial relations $E$. Each node is defined as $n_i = (c_i, s_i)$, where $c_i$ represents the node class, $s_i$ denotes the node state.

\textbf{Action level}: The action level analyzes human behaviors in the video and generates action sequences. Previous methods typically predict action sequences with custom models, requiring extensive data collection and labeling, and constraining the model's generalization capability. Although GPT-4V \cite{wake2023gpt} can be utilized to generate action sequences, its limited ability to assess the relative positions of objects hinders its overall accuracy. In contrast, we take full advantage of our hierarchical structure, employing LLMs to predict action sequences based on the observation level. This approach effectively captures relative positional relationships via scene graphs, thus improving the accuracy of video analysis.

Considering the redundancy in the constructed scene graphs for each downsampled frame, we select keyframes to reduce redundancy and facilitate action sequence generation. A keyframe is chosen when the scene graph of the current frame, denoted as $\bm{G}_t$, differs from that of the preceding frame, $\bm{G}_{t-1}$. This selection results in keyframe sequences, represented as $\bm{G}^k$, which have a length of $T^k$. For each keyframe, we convert the scene graph into text with the following format. 
\begin{tcolorbox}
[colback=gray!5!white,colframe=gray!75!black, left = 1mm, right = 1mm, top = 1mm, bottom = 1mm, sharp corners]
\small
Objects and states: \textbf{\textcolor[RGB]{106, 167, 78}{object1 [state], object2, object3 [state] ...}}

Inter-object relations: \textbf{\textcolor{blue}{object1 is [spatial relation] object2 ...}}

Human-object relations: \textbf{\textcolor[RGB]{74, 134, 231}{object3 is [spatial relation] hand.}}
\end{tcolorbox}
These keyframes are fed into LLMs to generate action sequences $\bm{\mathcal{A}}$ with a length of $T^{k}-1$ through the comparison between consecutive keyframes. More concretely, the $i-th$ action $\bm{\mathcal{A}}_{i}$ is derived as follows:
\begin{tcolorbox}
[colback=gray!5!white,colframe=gray!75!black, left = 1mm, right = 1mm, top = 1mm, bottom = 1mm, sharp corners]
\small
The scene graph of the first frame is  \textbf{\textcolor{blue}{$\bm{G}^{k}_i$}}, the scene graph of the second frame is  \textbf{\textcolor{blue}{$\bm{G}^{k}_{i+1}$}}

\textbf{Q}: What action did this human perform? \sethlcolor{yellow}\hl{\textbf{A}: Put chips in the drawer.}
\end{tcolorbox}

Consequently, the action knowledge comprises action sequences along with their corresponding initial and end scene graphs.
To facilitate summarization while reducing computational and memory resources, only objects pertinent to associated actions and those with spatial relations to the action-relevant objects are considered in the resulting scene graphs.

\textbf{Pattern level}: 
To effectively generalize in unseen  scenarios, we further distill the action level into the pattern level by summarizing
patterns presented in human demonstration videos. Furthermore, we independently analyze object-related and task-related patterns, which facilitates the retrieval of relevant knowledge for specific objects or tasks within unseen  scenarios. 
To achieve this, the generated action sequences and their associated scene graphs from the action knowledge, are fed into the LLM, and we instruct it to summarize these patterns.
\begin{tcolorbox}
[colback=gray!5!white,colframe=gray!75!black, left = 1mm, right = 1mm, top = 1mm, bottom = 1mm, sharp corners]
\small

Initial scene graph: \textcolor{blue}{$\bm{G}^{k}_0$}

Action and resulting scene graphs: 1. \textcolor[RGB]{246, 185, 76}{$\bm{\mathcal{A}}_0$}; \textcolor{blue}{$\bm{G}^{k}_{1}$}. 2. \textcolor[RGB]{246, 185, 76}{$\bm{\mathcal{A}}_1$}; \textcolor{blue}{$\bm{G}^{k}_{2}$}...

\textbf{Q}: 1. Concretize this task. 2. Summarize patterns for each object. \sethlcolor{yellow}\hl{\textbf{A}: 1. Repositioning each object on the table to its original location. 2. The drawer on the table
tends to contain snacks ...}
\end{tcolorbox}
In our implementation, we directly query the LLM to summarize task patterns $\bm{\mathcal{P}}^{t}$ and object patterns $\bm{\mathcal{P}}^{o}$. Specifically, the patterns related to tasks primarily involve concretizing human-provided instructions, enhancing the robot's comprehension of abstract human instructions. 
Additionally, we capture patterns for each object relevant to actions. For objects with single interactions,  their patterns are directly represented by the scene graph depicted in the keyframe. For objects involved in multiple interactions, we guide the LLM to summarize their patterns based on the associated actions and corresponding scene graphs. 

\textbf{Knowledge storage}: Following the acquisition of generalizable knowledge, we employ the summarized task patterns from the provided video and the visual representation of the scene as text-keys $\bm{K}^t$ and visual-keys $\bm{K}^v$, respectively. The acquired knowledge is stored in a knowledge base $\bm{\mathcal{B}}$ associated with these keys. The task patterns are stored in textual format, while the visual representations are preserved as images.

\subsection{Knowledge Retrieval} \label{retrieval}
When the robot operates in response to human instructions. DigKnow queries the knowledge base to retrieve pertinent stored knowledge, which then guides its planning process.
We observe that object-related knowledge within similar scenes can be transferrable across distinct tasks (e.g., object locations). Additionally, task-related knowledge for similar instructions can also be applicable across various scenes. 

The task knowledge corresponding to relevant instructions and the object knowledge corresponding to identical objects within similar scenes are extracted. 
Specifically, DigKnow initiates a zero-shot query to the LLM, instructing it to identify all tasks with text-keys $\bm{K}^t$ that exhibit semantic similarity to the current task instruction. The task pattern knowledge from each relevant task is incorporated into the planning process.
\begin{tcolorbox}
[colback=gray!5!white,colframe=gray!75!black, left = 1mm, right = 1mm, top = 1mm, bottom = 1mm, sharp corners]
\small
Current task: \textbf{\textcolor[RGB]{159, 140, 118}{[task instruction]}}

Previous tasks: 1. \textcolor[RGB]{74, 134, 231}{$\bm{K}^t_1$}, 2. \textcolor[RGB]{74, 134, 231}{$\bm{K}^t_2$}, 3. \textcolor[RGB]{74, 134, 231}{$\bm{K}^t_3$} ...

\textbf{Q}: 1. Does the new task belong to the same category as any previous tasks?  2. If it does, which specific prior tasks fall into this category? \sethlcolor{yellow}\hl{\textbf{A}: 1: Yes, 2: [1,5]}.
\end{tcolorbox}
Subsequently, we identify object instances within the current scene and access the pertinent object pattern knowledge stored in the knowledge base. Specifically, our approach calculates the visual similarity between the current scene and the visual-keys $\bm{K}^v$ stored in the knowledge base, selecting the top $N$ scenes displaying the highest similarity. Then, the pattern knowledge $\bm{\mathcal{P}}^{o}$ of the detected objects is retrieved from the selected scenes and transmitted to the LLM. In this paper, we employ DINO-V2 \cite{oquab2023dinov2} features for visual-visual retrieval.

Utilizing the retrieved pattern knowledge $\bm{\mathcal{P}}^{t}$ and $\bm{\mathcal{P}}^{o}$ for both task and object instances. DigKnow constructs the initial scene graph $\bm{G}_{0}$ of the current scene, and then LLM performs planning to generate the necessary action sequences $\bm{\hat{\mathcal{A}}}$ for task completion as follows:
\begin{tcolorbox}
[colback=gray!5!white,colframe=gray!75!black, left = 1mm, right = 1mm, top = 1mm, bottom = 1mm, sharp corners]
\small
Task: \textbf{\textcolor[RGB]{159, 140, 118}{[task instruction]}}

Initial scene graph: \textcolor{blue}{$\bm{G}_0$}

Pattern knowledge: \textcolor[RGB]{178, 123, 155}{$\bm{\mathcal{P}}^{t}$}, \textcolor[RGB]{178, 123, 155}{$\bm{\mathcal{P}}^{o}$}

\textbf{Q}: Please generate an action sequence to complete the task. \sethlcolor{yellow}\hl{\textbf{A}: 1. Open the drawer; 2. Pick up the drink; 3. Put the drink on the table; 4. Close the drawer.}
\end{tcolorbox}

\subsection{Knowledge-based Correction}
While the LLM generates plans informed by the retrieved knowledge, the generalization capacity remains constrained by the inherent instability in LLM outputs and action execution. To address this issue, the planning corrector and execution corrector are developed to leverage hierarchical knowledge for the validation and rectification of planning and execution outcomes. This approach significantly improves task completion success rates and enhances generalization capabilities.

\textbf{Planning corrector}: We observe that planning failures can predominantly be attributed to two factors: (1) inconsistencies between plans and the provided knowledge, and (2) planning actions without fulfilling their preconditions, exemplified by instances where a robot intends to close the cabinet before placing the drink it holds in its gripper.
To address the first factor, we query LLM to assess the alignment between generated plans and the retrieved knowledge, resulting in the summarization of disparities denoted as $\bm{\mathcal{S}}^k$. 
\begin{tcolorbox}
[colback=gray!5!white,colframe=gray!75!black, left = 1mm, right = 1mm, top = 1mm, bottom = 1mm, sharp corners]
\small
Task: \textbf{\textcolor[RGB]{159, 140, 118}{[task instruction]}}

Pattern knowledge: \textcolor[RGB]{178, 123, 155}{$\bm{\mathcal{P}}^{t}$}, \textcolor[RGB]{178, 123, 155}{$\bm{\mathcal{P}}^{o}$}

Planed actions: \textcolor[RGB]{246, 185, 76}{$\bm{\hat{\mathcal{A}}}_0$}, \textcolor[RGB]{246, 185, 76}{$\bm{\hat{\mathcal{A}}}_1$}, ...

\textbf{Q}: 1. Validate compliance with pattern knowledge; 2. if not, summarize accordingly. \sethlcolor{yellow}\hl{\textbf{A}: 1. No. 2. The drink should be in the drawer, not in the basket. ...}
\end{tcolorbox}
Regarding the second factor, our approach involves three key steps. Initially, we instruct LLM to infer the preceding scene graphs $\hat{\bm{G}}^{k}_{t}$ for individual actions $\bm{\hat{\mathcal{A}}}_t$ by leveraging the initial scene graph and the planned action sequence. Subsequently, actions $\bm{{\mathcal{A}}}_t$ and their preceding scene graphs ${\bm{G}}^{k}_{t}$ from retrieved knowledge are introduced (Sec. \ref{retrieval}). Then, LLM assesses the generated action sequence $\bm{\hat{\mathcal{A}}}$ for compliance with execution requirements by considering the provided actions $\bm{{\mathcal{A}}}_t$ and their associated preceding scene graphs ${\bm{G}}^{k}_{t}$, while also providing summaries of any discrepancies and their explanations $\bm{\mathcal{S}}^c$.
\begin{tcolorbox}
[colback=gray!5!white,colframe=gray!75!black, left = 1mm, right = 1mm, top = 1mm, bottom = 1mm, sharp corners]
\small
Action and preceding scene graphs: 1. \textcolor[RGB]{246, 185, 76}{$\bm{\mathcal{A}}_0$}; \textcolor{blue}{$\bm{G}^{k}_{0}$}. 2. \textcolor[RGB]{246, 185, 76}{$\bm{\mathcal{A}}_1$}; \textcolor{blue}{$\bm{G}^{k}_{1}$}...

Initial scene graph: \textcolor{blue}{$\bm{G}_0$}

Planed actions: \textcolor[RGB]{246, 185, 76}{$\bm{\hat{\mathcal{A}}}_0$}, \textcolor[RGB]{246, 185, 76}{$\bm{\hat{\mathcal{A}}}_1$}, ...

\textbf{Q}: 1. Infer the scene graph sequence based on the planned action and the initial scene graph. 2. Determine compliance with execution conditions based on the provided action, its preceding scene graph, and the inferred preceding scene graph 3. If not compliant, summarize the discrepancies. \sethlcolor{yellow}\hl{\textbf{A}: 1. $\hat{\bm{G}}^{k}_{0}$, $\hat{\bm{G}}^{k}_{1}$, $\hat{\bm{G}}^{k}_{2}$... 2. No. 3. The robot should first place the drink and then close the drawer..}
\end{tcolorbox}
Based on the explanation for planning failures summarized by LLM, we instruct LLM to make corrections to the plans.

\begin{tcolorbox}
[colback=gray!5!white,colframe=gray!75!black, left = 1mm, right = 1mm, top = 1mm, bottom = 1mm, sharp corners]
\small
Planed actions: \textcolor[RGB]{246, 185, 76}{$\bm{\hat{\mathcal{A}}}_0$}, \textcolor[RGB]{246, 185, 76}{$\bm{\hat{\mathcal{A}}}_1$}, ...

Failure explanation: \textcolor{pink}{$\bm{\mathcal{S}}^k$}, \textcolor{pink}{$\bm{\mathcal{S}}^c$}

\textbf{Q}: Revise plans based on failure explanation. \sethlcolor{yellow}\hl{\textbf{A}: 3. Place the drink on the table. 4. Close the drawer ...}
\end{tcolorbox}

\textbf{Execution corrector}: To alleviate task failure caused by individual action execution errors, we employ an execution correction mechanism that leverages retrieved knowledge for validation. Following the execution of each action, we construct a scene graph of the current environment.  To refine this scene graph, we consider only objects relevant to the action and those spatially associated with task-relevant objects. Subsequently, semantically analogous actions and their resulting scene graphs are introduced. The LLM then evaluates whether the execution outcomes of the current action align with those of the provided actions. Disparities and their explanations are summarized. Based on this summary, we instruct LLM to rectify the failure.
\begin{tcolorbox}
[colback=gray!5!white,colframe=gray!75!black, left = 1mm, right = 1mm, top = 1mm, bottom = 1mm, sharp corners]
\small
Action and resulting scene graphs: 1. \textcolor[RGB]{246, 185, 76}{$\bm{\mathcal{A}}_0$}; \textcolor{blue}{$\bm{G}^{k}_{1}$}. 2. \textcolor[RGB]{246, 185, 76}{$\bm{\mathcal{A}}_1$}; \textcolor{blue}{$\bm{G}^{k}_{2}$}...

Executed action and the resulting scene graph: \textcolor[RGB]{246, 185, 76}{$\bm{\hat{\mathcal{A}}}_t$}, \textcolor{blue}{$\bm{G}_{t+1}$}

\textbf{Q}: 1. Based on the provided action and its corresponding resulting scene graph, whether the executed action is successful. 2. If not, explain the failure and generate the correction plan. \sethlcolor{yellow}\hl{\textbf{A}: 1. No. 2. Explanation: Failed to successfully grasp the drink. Correction plan: 1. Pick up the drink from the drawer; 2. Place the drink on the table.}
\end{tcolorbox}

\section{Experiments}
We evaluate our approach in a real-world environment with a Franka Emika Panda Robot. We use GPT-4 \cite{openai2023gpt4} for all LLM modules. We design qualitative and quantitative evaluation experiments to test DigKnow’s core capabilities: (\uppercase\expandafter{\romannumeral1})  distilling hierarchical knowledge from human demonstration videos, (\uppercase\expandafter{\romannumeral2})  retrieving relevant knowledge and performing planning, and (\uppercase\expandafter{\romannumeral3})  validating and correcting the planning and execution results. 

\subsection{Qualitative experiments}

\subsubsection{Generalizable knowledge distillation}

\begin{figure}[!h]
\setlength{\abovecaptionskip}{-0pt}
    \centering
    \includegraphics[width=0.45\textwidth]{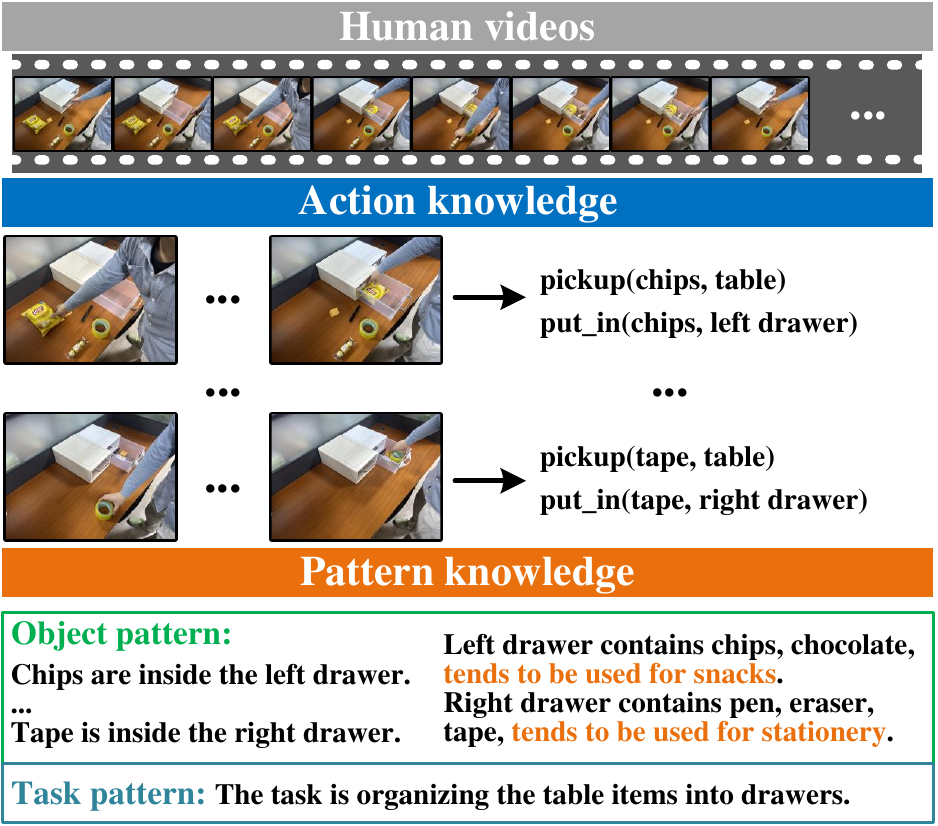}
    \caption{Qualitative results of generalizable knowledge distillation.}
    \label{fig3}
    \end{figure}

Effective knowledge distillation from videos is crucial for enhancing the adaptability of robots in diverse environments. To demonstrate the effect of generalizable knowledge distillation, we utilize a human demonstration video as input, and the resulting knowledge is presented in Fig \ref{fig3}. DigKnow effectively generates action sequences precisely, and distills task and object pattern knowledge, enabling our method to generalize to unseen scenarios.

\subsubsection{Knowledge-based planning}
\begin{figure}[!h]
\setlength{\abovecaptionskip}{0pt}
    \centering
    \includegraphics[width=8.5cm]{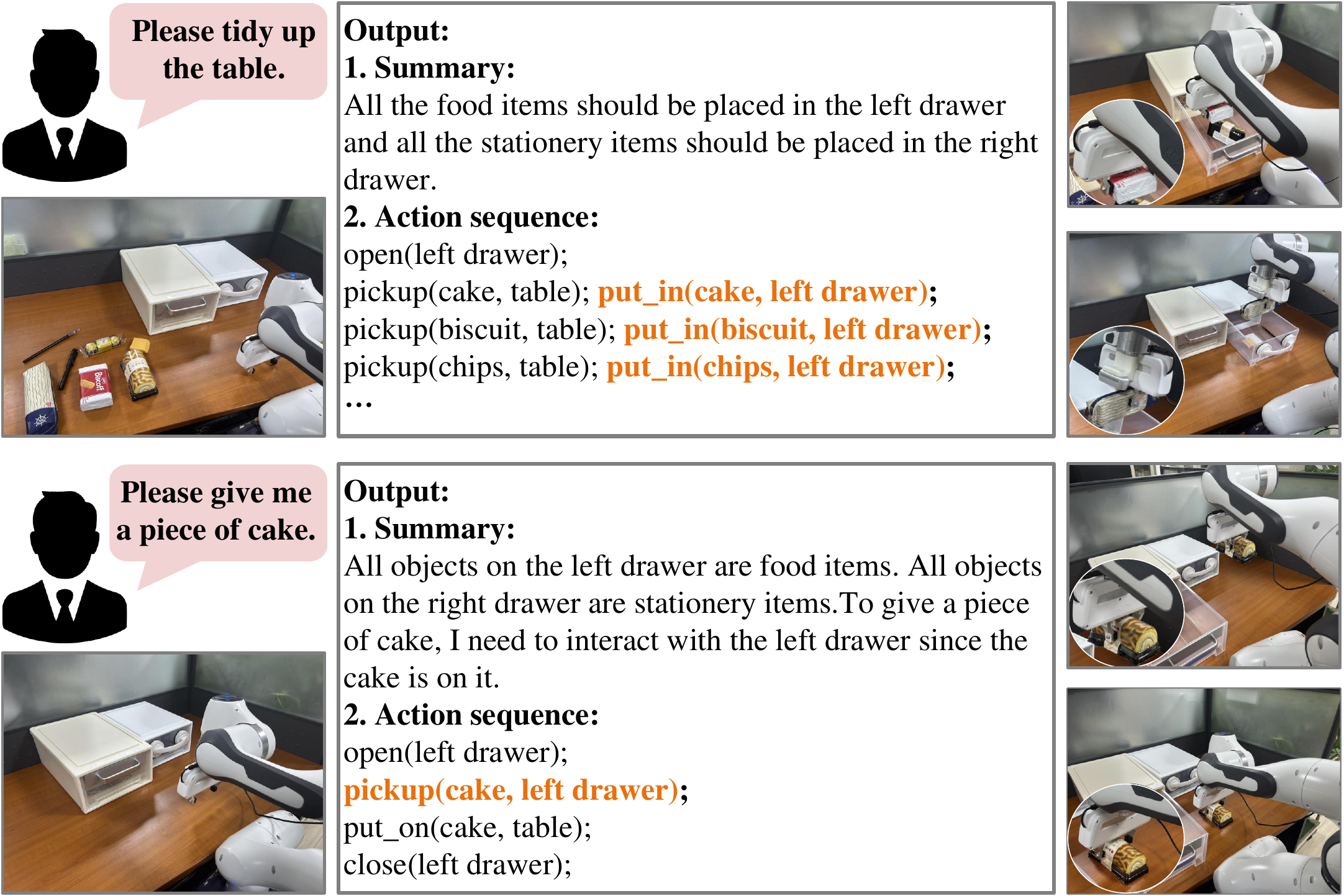}
    \caption{Qualitative results of knowledge-based planning. }
    \label{fig4}
    \end{figure}

To assess DigKnow's generalization capability through knowledge acquisition, we devise two scenarios featuring distinct object instances and task directives. We apply the acquired knowledge for planning, and the results, including a summary and the generated action sequence, are presented in Figure \ref{fig4}. In the first scenario, which involves the desk organization task with unseen object instances, DigKnow demonstrates precise task comprehension facilitated by distilled task pattern knowledge. 
In the second scenario, involving a different task and unseen object instances, DigKnow also achieves successful task execution.

\subsubsection{Knowledge-based correction}
\begin{figure}[!h]
\setlength{\abovecaptionskip}{5pt}
    \centering
    \includegraphics[width=8.5cm]{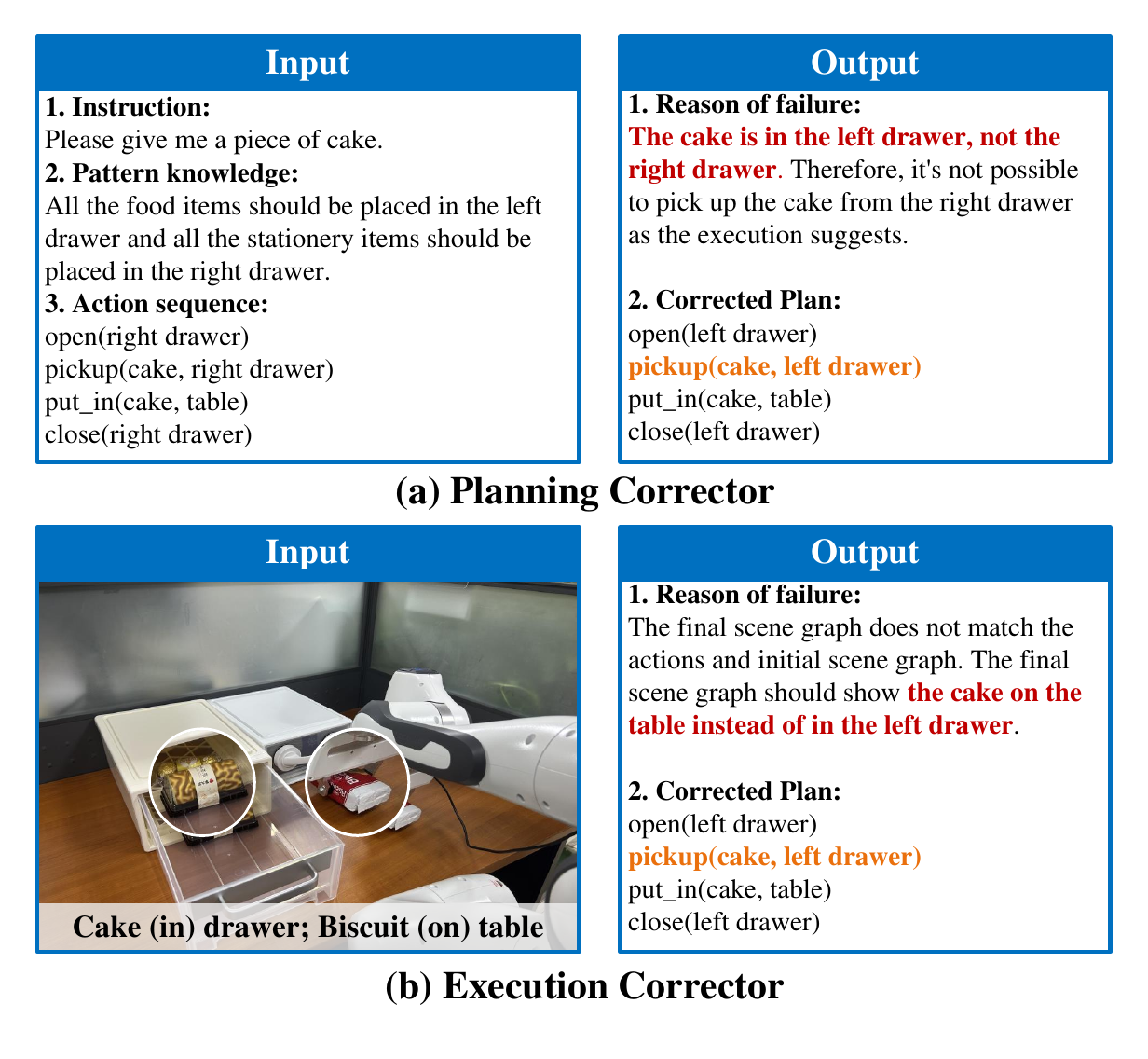}
    \caption{Qualitative results of planning corrector and execution corrector.}
    \label{fig5}
    \vspace{-10pt}
    \end{figure}

Verification and correction are crucial for enhancing the robot's generalization across different scenarios, as planning and execution errors are inevitable. To validate DigKnow's verification and correction capabilities based on extracted knowledge, we conduct qualitative investigations shown in Fig. \ref{fig5}. DigKnow is capable of correcting the planned action sequences by comparing them with the provided knowledge. Additionally, in cases of policy execution errors, DigKnow accurately analyzes the causes of the error and generates plans to rectify it.

\subsection{Quantitative experiments}

\subsubsection{Experiments and Tasks}
\begin{figure}[!h]
\setlength{\abovecaptionskip}{0pt}
    \centering
    \includegraphics[width=0.5\textwidth]{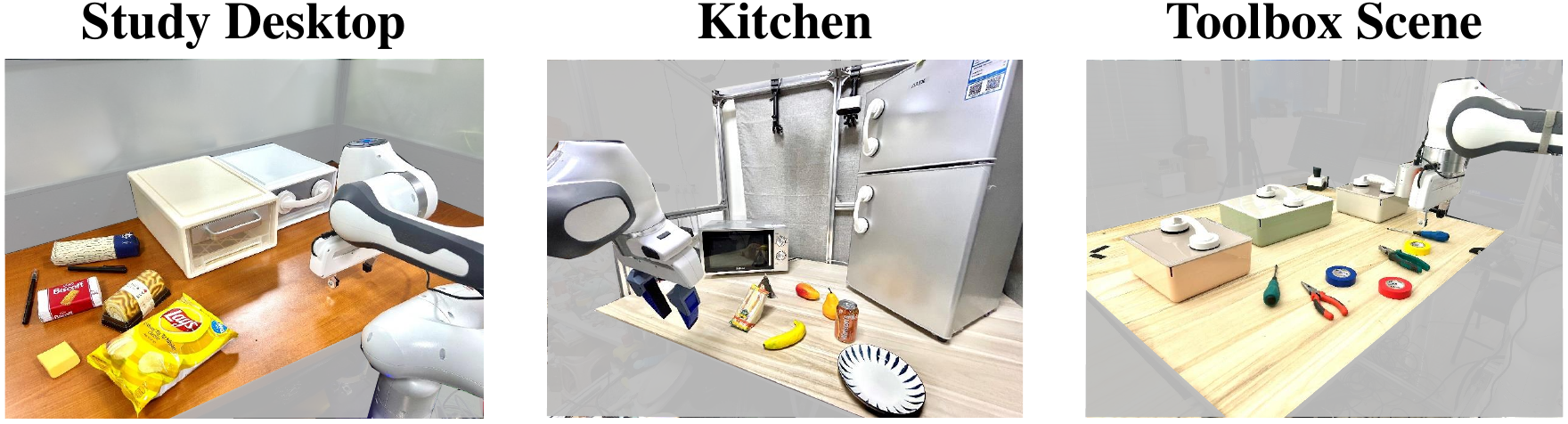}
    \caption{Three types of experimental environments.}
    \label{exp_setup}
    \end{figure}

We create three types of experimental environments (\textbf{Study Desktop}, \textbf{Kitchen}, and \textbf{Toolbox Scene}) encompassing 9 demonstration tasks (Seen Task) and 6 generalization tasks (Unseen Task) with different difficulty, focusing on long-horizon tasks with different goals, as shown in Fig. \ref{exp_setup}. 

The demonstration videos align with the object instances and task objectives in the Seen Tasks. For each Seen Task, five human demonstrations with variations in background and viewpoint were provided. In Seen Task experiments, as shown in Table. \ref{task_list_seen}, all methods learn human actions exclusively from their corresponding demonstrations. For Unseen Tasks, as shown in Table. \ref{task_list_unseen}, methods generate action sequences by leveraging learned action patterns from all demonstration videos, conditioned on the provided initial state and task objective.

\begin{table}[h!]
  \caption{Seen Task list.}
  \centering
  \resizebox{0.50\textwidth}{!}{
      \begin{tabular}{clcc}
      \toprule
      Seen task & Task description & Step & Type \\
      \midrule
      Task-1 & Put pen in right open drawer and close it & \cellcolor{green!50} 4 &  Destop \\ 
      Task-2 & Put cookie in left drawer and close it & \cellcolor{yellow!90} 6 &  Destop \\ 
      Task-3 & Put cake in left drawer and put eraser in right drawer & \cellcolor{red!50} 12 &  Destop \\ 
      Task-4 & Put stationery in right drawer and snack in left drawer & \cellcolor{red!50} 16 &  Destop \\ 
      \midrule
      Task-5 & Put fruit in plate & \cellcolor{green!50} 2 &  Kitchen \\ 
      Task-6 & Put sandwich in microwave and close it & \cellcolor{yellow!90} 6 &  Kitchen \\ 
      Task-7 & Put fruits in fridge and sandwich in microwave & \cellcolor{red!50} 10 &  Kitchen \\ 
      \midrule
      Task-8 & Put screwdriver in right box and close it & \cellcolor{yellow!90} 8 &  Toolbox \\ 
      Task-9 & Put tapes in left box and pincers in front box & \cellcolor{red!50} 14 &  Toolbox \\ 
      \bottomrule
      \end{tabular}
    }
  \vspace{-10pt}
  \label{task_list_seen}
\end{table}

\begin{table}[h!]
  \caption{Unseen Task list.}
  \centering
  \resizebox{0.50\textwidth}{!}{
      \begin{tabular}{clcc}
      \toprule
      Unseen task & Task description & Step & Type \\
      \midrule
      Task-10 & Please tidy up the table & \cellcolor{red!50} 12 &  Object generalization \\ 
      Task-11 & Please give me a cake and a pen & \cellcolor{red!50} 12 &  Task generalization \\ 
      \midrule
      Task-12 & Please help me heat breakfast & \cellcolor{red!50} 10 &   Object generalization \\ 
      Task-13 & Please give me a banana and sandwich & \cellcolor{red!50} 12 &  Task generalization \\ 
      \midrule
      Task-14 & Please help me organize the toolbox &  \cellcolor{red!50} 12 &   Object generalization \\ 
      Task-15 & Please give me a screwdriver and a tape &  \cellcolor{red!50} 12 &  Task generalization \\ 
      \bottomrule
      \end{tabular}
    }
  \label{task_list_unseen}
\end{table}

\begin{table*}[ht!]
  \caption{Quantitative results of generalizable knowledge distillation.}
  \vspace{-0mm}
  \centering
  \resizebox{0.95\textwidth}{!}{
      \begin{tabular}{lcccccccccc}
      \toprule
      \multirow{2}{*}{\textbf{Methods}} & \multicolumn{4}{c}{\textbf{Study Destop}} & \multicolumn{3}{c}{\textbf{Kitchen}} & \multicolumn{2}{c}{\textbf{Toolbox Scene}} & \multirow{2}{*}{\begin{tabular}[c]{@{}c@{}}\textbf{Average} \\ \textbf{Score}\end{tabular}} \\
      \cmidrule(lr){2-5}
      \cmidrule(lr){6-8} 
      \cmidrule(lr){9-10}
       & \cellcolor{green!50}Task-1 & \cellcolor{yellow!90} Task-2 & \cellcolor{red!50} Task-3 &\cellcolor{red!50} Task-4 &\cellcolor{green!50} Task-5 &\cellcolor{yellow!90} Task-6 &\cellcolor{red!50} Task-7 & \cellcolor{yellow!90} Task-8&
       \cellcolor{red!50} Task-9 &  \\
      \midrule
GPT-4V &             0.67 & 0.39 & 0.01 & 0.00    & 0.91 & 0.52 & 0.04  & 0.41 & 0.19 & 0.35 \\
Demo2Code &          0.69 & 0.30 & 0.00 & 0.01    & 0.94 & 0.46 & 0.01  & 0.37 & 0.12 & 0.32 \\
GPT4V-Robo &        0.75 & 0.41 & 0.16 & 0.09    & 0.97 & 0.59 & 0.19  & 0.67 & 0.22 & 0.45 \\
GPT4V-Robo-l &      0.88 & 0.68 & 0.48 & 0.41    & 0.92 & 0.79 & 0.47  & 0.71 & 0.45 & 0.64 \\
\rowcolor{gray!25} 
DigKnow &            \textbf{0.92} & \textbf{0.83} & \textbf{0.72} & \textbf{0.69}   & \textbf{0.97} & \textbf{0.86} & \textbf{0.63}  & \textbf{0.89} & \textbf{0.79}&\textbf{0.81} \\
      \bottomrule
      \end{tabular}
  }
  \vspace{-3mm}
\label{EXP_1_TABLE}
\end{table*}

\subsubsection{Baselines and Metrics}
DigKnow is compared against four representative methods: (1) Standard GPT-4V \cite{openai2023gpt4}, a state-of-the-art vision-language model; (2) Demo2Code \cite{wang2023demo2code}, an LLM-based planner that converts demonstrations into executable task code (adapted here to integrate GPT-4V’s vision analysis capabilities); (3) GPT4V-Robo \cite{wake2023gpt4v}, a robotics-focused variant of GPT-4V optimized for one-shot visual instruction in manipulation tasks; and (4) GPT4V-Robo-l, an extended version of GPT4V-Robo \cite{wake2023gpt4v} modified to process long-horizon human demonstration videos. 
All baseline methods output sequences of action primitives (following RoboCodeX \cite{mu2024robocodex}). For real-robot experiments, we employed the calibrated RGB-D cameras to capture 3D observation and a ROS-based control framework to execute the planned action sequences on a Franka robot.

In motion planning experiments, following \cite{wake2023gpt4v}, we evaluate the quality of the output action plan against ground truth using the normalized Levenshtein distance, which ranges from 0 to 1, with 1 indicating a perfect match. For each task, experiments are conducted across five distinct initial states, each repeated ten times. The mean quality across all trials serves as the score. 
In real-robot experiments, we evaluate the average success rate of the task. For each task, experiments are conducted 20 times with different backgrounds and object positions.

\subsubsection{Motion planning experiments}
To evaluate DigKnow’s capabilities in parsing human demonstration videos and task planning, we conducted motion planning experiments on 9 Seen tasks and 6 Unseen tasks.
Quantitative results for Seen and Unseen tasks, presented in Table \ref{EXP_1_TABLE} and Table \ref{EXP_2_TABLE} respectively, obviously exhibit a substantial enhancement achieved by DigKnow over baseline methods. 
In Seen task experiments, DigKnow exhibit superior performance in comprehending demonstration videos and action planning, while baseline methods struggle to generate the correct action sequence, particularly in long-horizon tasks.
In Unseen task setups, DigKnow adapts robustly to variations in object instances and task objectives.
In contrast, baseline methods exhibit significant performance degradation in such scenarios. 

\begin{table}[h!]
\vspace{-5pt}
\caption{Quantitative results of knowledge-based planning.}
\centering
\centering
\resizebox{0.50\textwidth}{!}{
\begin{tabular}{lccccccc}
\toprule
\multirow{2}{*}{\textbf{Unseen task}} & \multicolumn{2}{c}{\textbf{Destop}} & \multicolumn{2}{c}{\textbf{Kitchen}} & \multicolumn{2}{c}{\textbf{Toolbox}} & \multirow{2}{*}{\textbf{Avg.}} \\
      \cmidrule(lr){2-3}
      \cmidrule(lr){4-5} 
      \cmidrule(lr){6-7}
       & \cellcolor{red!50} T-10 & \cellcolor{red!50} T-11 & \cellcolor{red!50} T-12 & \cellcolor{red!50} T-13 & \cellcolor{red!50} T-14 & \cellcolor{red!50} T-15  \\
      \midrule
      
GPT-4V &               0.10 & 0.12 & 0.35 & 0.11 & 0.09 & 0.29 & 0.18\\
Demo2Code &            0.02 & 0.13 & 0.27 & 0.07 & 0.04 & 0.31 & 0.14 \\
GPT4V-Robo &           0.27 & 0.46 & 0.46 & 0.23 & 0.19 & 0.33 & 0.32 \\
GPT4V-Robo-l &       0.51 & 0.65 & 0.57 & 0.55 & 0.58 & 0.45 & 0.55 \\
\rowcolor{gray!25}  
\textbf{DigKnow} &             \textbf{0.79 }&\textbf{0.85 }& \textbf{0.82}& \textbf{0.77} & \textbf{0.71} & \textbf{0.85} & \textbf{0.80} \\
\bottomrule
\end{tabular}
}
\vspace{-5pt}
\label{EXP_2_TABLE}
\end{table}

\subsubsection{Real-robot experiments}
we conduct the real-robot experiments to evaluate the performance of all methods in both seen and unseen tasks.  Quantitative results, presented in Table \ref{real_robot_table}, indicate that DigKnow achieves superior performance in real-world scenarios compared to baseline methods, particularly in long-horizon tasks and Unseen Tasks.
This advantage is attributed to the hierarchical knowledge base and the correction modules. 

\begin{table*}[ht!]
  \caption{Quantitative results of real-robot experiments.}
  \vspace{-2mm}
  \centering
  \resizebox{1.00\textwidth}{!}{
      \begin{tabular}{lcccccccccccccccc}
      \toprule
      \multirow{2}{*}{\textbf{Methods}} & \multicolumn{4}{c}{\textbf{Destop}} & \multicolumn{3}{c}{\textbf{Kitchen}} & \multicolumn{2}{c}{\textbf{Toolbox}} & \multicolumn{2}{c}{\textbf{Destop}} & \multicolumn{2}{c}{\textbf{Kitchen}} & \multicolumn{2}{c}{\textbf{Toolbox}} & \multirow{2}{*}{Avg.} \\
      \cmidrule(lr){2-5}
      \cmidrule(lr){6-8} 
      \cmidrule(lr){9-10}
      \cmidrule(lr){11-12}
      \cmidrule(lr){13-14}
      \cmidrule(lr){15-16}
       & \cellcolor{green!50}T-1 & \cellcolor{yellow!90} T-2 & \cellcolor{red!50} T-3 &\cellcolor{red!50} T-4 &\cellcolor{green!50} T-5 &\cellcolor{yellow!90} T-6 &\cellcolor{red!50} T-7 & \cellcolor{yellow!90} T-8& \cellcolor{red!50} T-9 & \cellcolor{red!50} T-10 & \cellcolor{red!50} T-11 & \cellcolor{red!50} T-12 & \cellcolor{red!50} T-13 & \cellcolor{red!50} T-14 & \cellcolor{red!50} T-15  \\
      \midrule
GPT4V-Robo &        0.70& 0.35 & 0.05 & 0.00    & 0.85 & 0.25 & 0.00    & 0.30 & 0.05  & 0.10 & 0.25 & 0.30 & 0.05 & 0.10 & 0.20 & 0.24 \\
GPT4V-Robo-l &      0.65 & 0.55 & 0.35 & 0.30    & 0.80 & 0.55 & 0.25   & 0.45 & 0.30 & 0.35 & 0.45 & 0.55 & 0.35 & 0.45 & 0.35 & 0.45\\
\rowcolor{gray!25}
\textbf{DigKnow} &           \textbf{0.80} &\textbf{0.65} & \textbf{0.65} & \textbf{0.60}    & \textbf{0.90} & \textbf{0.75} & \textbf{0.50}  & \textbf{0.75} & \textbf{0.60} & \textbf{0.65} & \textbf{0.70}& \textbf{0.65}& \textbf{0.60} & \textbf{0.55} & \textbf{0.70} & \textbf{0.73}\\
      \bottomrule
      \end{tabular}
  }
  \vspace{-3mm}
\label{real_robot_table}
\end{table*}

\subsection{Ablation study}
\begin{figure}[!h]
\setlength{\abovecaptionskip}{-5pt}
    \centering
    \includegraphics[width=0.5\textwidth]{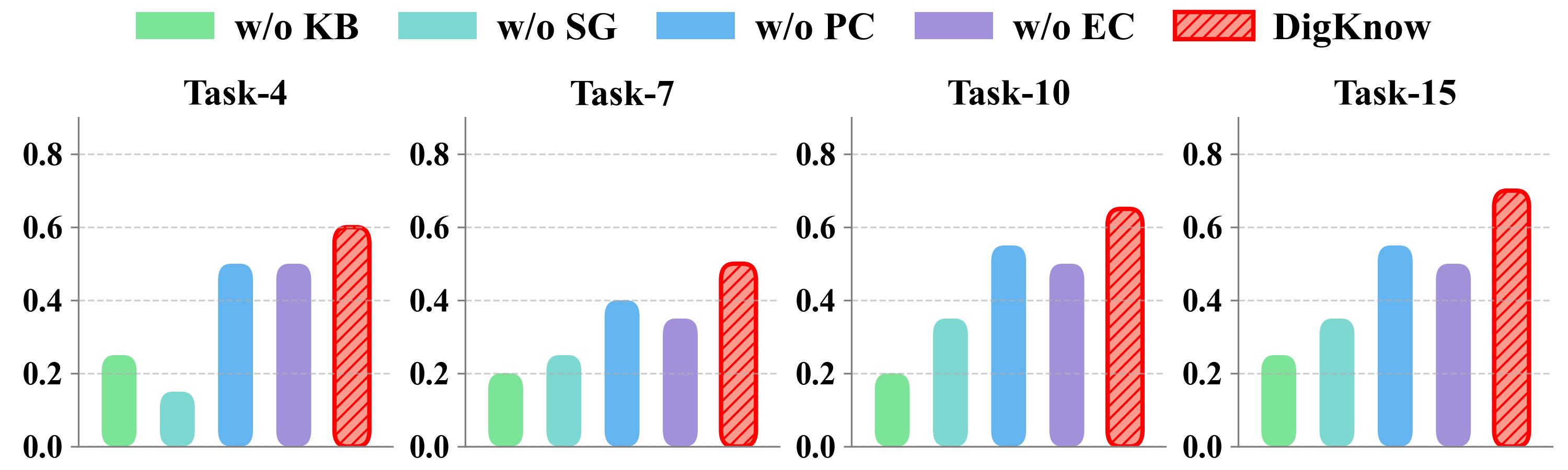}
    \caption{Quantitative results of ablation study.}
    \label{ablation}
    \vspace{-5pt}
    \end{figure}

Comprehensive ablation studies are conducted on four representative real-robot tasks to investigate the fundamental designs of DigKnow, which focus on examining the utility of \textbf{Knowledge base}, \textbf{Scene graph representation}, \textbf{Planning correction} and \textbf{Execution correction} module. Experimental results, as shown in Fig. \ref{ablation}, demonstrate the effectiveness of each module designed by Digknow.
Specifically, (1) \textbf{Knowledge Base (KB)}, which is the core contribution of DigKnow, effectively enhances the performance through distilling and summarizing the hierarchical knowledge in human video; (2) \textbf{Scene Graph (SG)} representation mitigates hallucination issues in object recognition and spatial localization for Vision-Language Models, improves overall task performance; (3) \textbf{Planning Correction (PC)} module effectively improve the performance of motion planning, particularly in long-horizon tasks and Unseen tasks;
(4) \textbf{Execution Correction (EC)} module precisely diagnoses execution errors and generates corrective plans, thereby improving task success rates in real-robot experiments.

\begin{figure}[!h]
\vspace{-5pt}
\setlength{\abovecaptionskip}{-5pt}
    \centering
    \includegraphics[width=0.5\textwidth]{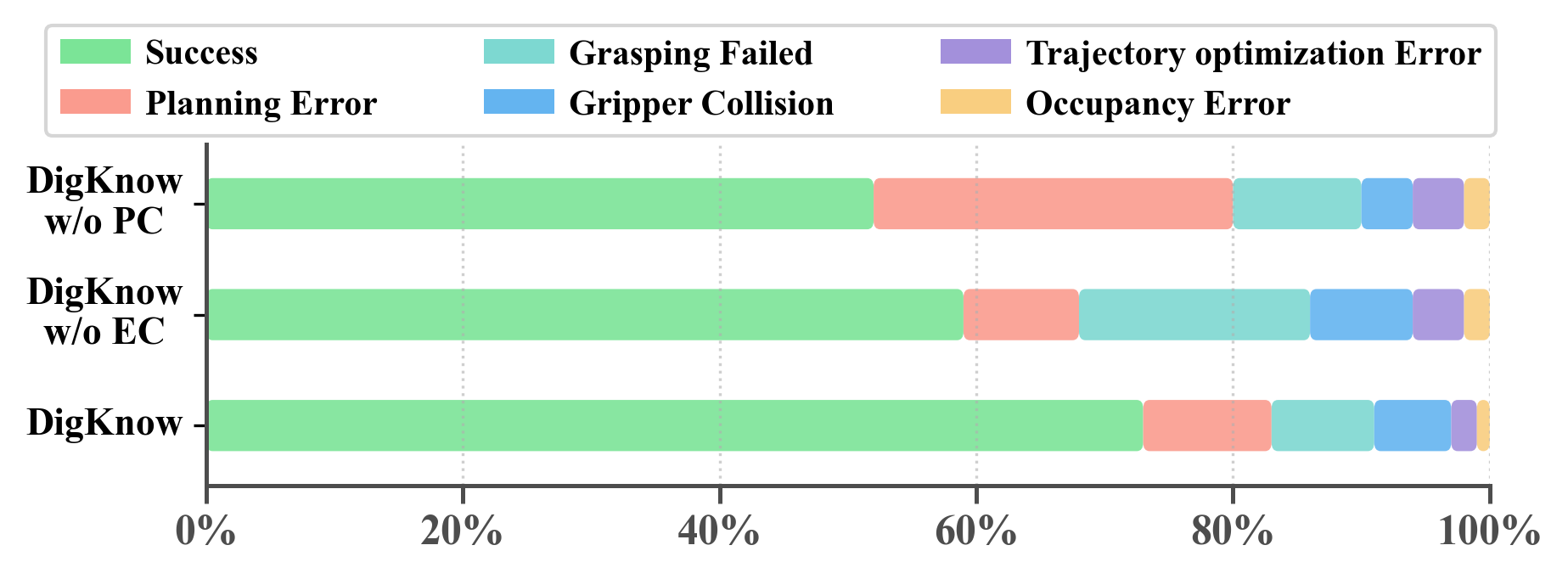}
    \caption{{Quantitative results of error analyses.} }
    \label{fig6}
    \end{figure}

We specifically conduct error analyses for real-robot experiments to evaluate the correction module. As shown in Fig. \ref{fig6}, the Planning correction module effectively mitigates task failures caused by flawed planning, while the Execution correction module iteratively refines plans based on observation data and retries failed subtasks, collectively enhancing overall system robustness.

\newpage

\section{Conclusion}
In this paper, we introduce a novel perspective regarding human videos as knowledge for robots, departing from the utilization of videos solely as instructions, as employed in prior methods. To achieve this, we introduce an efficient and robust framework, termed DigKnow, designed to distill generalizable knowledge from human videos, spanning observation, action, and pattern levels. In diverse task and object instance scenarios, DigKnow selectively retrieves pertinent knowledge for the present task and object instances.  Subsequently, our LLM-based planner formulates plans based on the retrieved knowledge, while the policy executes actions in accordance with the plan to achieve the specified task. Leveraging the retrieved generalizable knowledge, we validate and rectify planning and execution outcomes, resulting in a substantial improvement in the success rate.  Experimental results conducted across various tasks and environments demonstrate the effectiveness of this approach in enabling real-world robots to accomplish tasks with knowledge derived from human demonstrations.

{
\bibliographystyle{unsrt}

\bibliography{egbib}
}

\end{document}